\documentclass[runningheads]{llncs}

\usepackage{eccv}

\usepackage{eccvabbrv}

\usepackage{graphicx}
\usepackage{booktabs}
\usepackage[numbers,sort&compress]{natbib}
\usepackage{algorithm}
\usepackage{algpseudocode}
\usepackage{tabularx}
\usepackage{array}
\usepackage{makecell}
\usepackage{adjustbox}
\usepackage{multirow}
\usepackage{siunitx}
\usepackage[accsupp]{axessibility}  %

\usepackage[hidelinks]{hyperref}
\hypersetup{
    colorlinks,
    linkcolor={blue!80!black},
    citecolor={blue!80!black},
    urlcolor={blue!80!black}
}
\usepackage{orcidlink}

\newcommand{\method}{VISOR++}

\begin{document}

\title{VISOR++: Visual Input based Steering for Large Vision Language Models} 

\titlerunning{VISOR++}

\author{
Ravi Balakrishnan\thanks{These authors contributed equally.}\inst{1}\orcidlink{0000-0001-8490-8486}
\and
Mansi Phute$^\star$\inst{2}\orcidlink{0000-0003-4682-6281}
}

\authorrunning{R. Balakrishnan and M. Phute}
\institute{HiddenLayer, USA\\
\email{b.ravikumar88@gmail.com} \and
Georgia Institute of Technology, USA\\
\email{mansiphute@gatech.edu}\\
}

\maketitle

\begin{abstract}
As Vision Language Models (VLM) are deployed across safety-critical applications, understanding and controlling their behavioral patterns has become increasingly important. Existing behavioral control methods face significant limitations: system prompting is a popular approach but could easily be overridden by user instructions, while applying activation-based steering vectors requires invasive runtime access to model internals, precluding deployment with API-based services and closed-source models. Finding steering methods that transfer across multiple VLMs is still an open area of research. To this end, we introduce visual input based steering for output redirection (VISOR++), a novel approach that achieves behavioral control through optimized visual inputs alone. We demonstrate that a single VISOR++ image can be generated for two architecturally diverse VLMs that by itself can emulate each of their steering vectors. By crafting universal visual inputs that induce target activation patterns for an ensemble of models, VISOR++ eliminates the need for runtime model access while remaining deployment-agnostic. This means that when an underlying model supports multimodal capability, model behaviors can be steered by inserting an image input completely replacing runtime steering vector based interventions. We first demonstrate the effectiveness of the VISOR++ images on open-access models such as LLaVA-1.5-7B and IDEFICS2-8B along three alignment directions: refusal, sycophancy and survival instinct. Both the model-specific steering images and the jointly optimized images achieve performance parity closely following that of steering vectors for both positive and negative steering tasks. We also show early promise of VISOR++ images in achieving directional behavioral shifts for unseen models that include both open-access and closed-access models. At the same time, VISOR++ images exhibit no measurable degradation on 14,000 unrelated MMLU evaluation samples highlighting their specificity to inducing only behavioral shifts.
\end{abstract}

\section{Introduction}
\label{sec:intro}

Vision-Language Models (VLM) process both images and text to enable applications ranging from visual question answering and image captioning to multimodal reasoning and code generation from screenshots~\citep{achiam2024gpt4, touvron2023llama2}. These models are increasingly deployed in production systems, including safety-critical domains like healthcare, autonomous systems, and content moderation, where they at times outperform text-only models even on purely textual tasks due to their richer pre-training. As VLMs become core infrastructure for both multimodal and text-based applications, ensuring their behavioral alignment and resistance to adversarial manipulation becomes essential for preventing harmful outputs and maintaining system reliability.

Researchers have developed methods for bypassing alignment in Large Language Models (LLM), including prompt engineering \citep{liu2023jailbreaking}, adversarial suffixes \citep{zou2023universal}, and steering vectors \citep{turner2023activation, panickssery2023steering}. Numerous attacks targeting VLMs have been explored, including manipulation of image embeddings, adversarial patching, prompt injection, and inpainting techniques \citep{bailey2023image, qi2023visual, shayegani2023jailbreak}. Steering vectors, in particular, function by manipulating the activation space of a model. A popular steering technique involves computing the steering vector as the difference between the activations corresponding to the undesired and desired outputs. When added to the model's activation layers during inference, it induces targeted behavioral shifts. While powerful, the practical application of steering vectors is fundamentally constrained by their requirement for white-box access to model internals, including the need to compute and manipulate activations at runtime, an assumption that does not hold in many realistic settings.

The above limitation is significant since, on the one hand, inaccessibility of model internals in production systems creates a false sense of security against activation-based attacks. On the other hand, the applicability of guardrailing using steering becomes severely restricted since the majority of the VLMs are served via APIs without access to inference pipelines.

In order to make the steering techniques for VLMs practicable, we introduce \method{} (\textbf{V}isual \textbf{I}nput based \textbf{S}teering for \textbf{O}utput \textbf{R}edirection), a technique that optimizes perturbations in the input image space to mimic the behavior of steering vectors in the latent activation space. We successfully demonstrate the existence of images that can steer model behavior across a range of input text prompts for three different behavioral dimensions. We show that both per-model steering images as well as a single image trained across two different models can achieve similar levels of steering as their corresponding steering vectors in most cases. We show the effects of the jointly trained image on unseen models. While we do not yet fully observe a strong transfer, we highlight that directional transfer (especially to induce negative behavior) is possible even when trained over a limited ensemble size of 2. We believe our findings provide interesting insights towards understanding the relationship between visual inputs and model hidden states and helps take a firm step towards developing truly transferable behavioral steering images.

\begin{figure}[t]
\centering
\includegraphics[width=\textwidth]{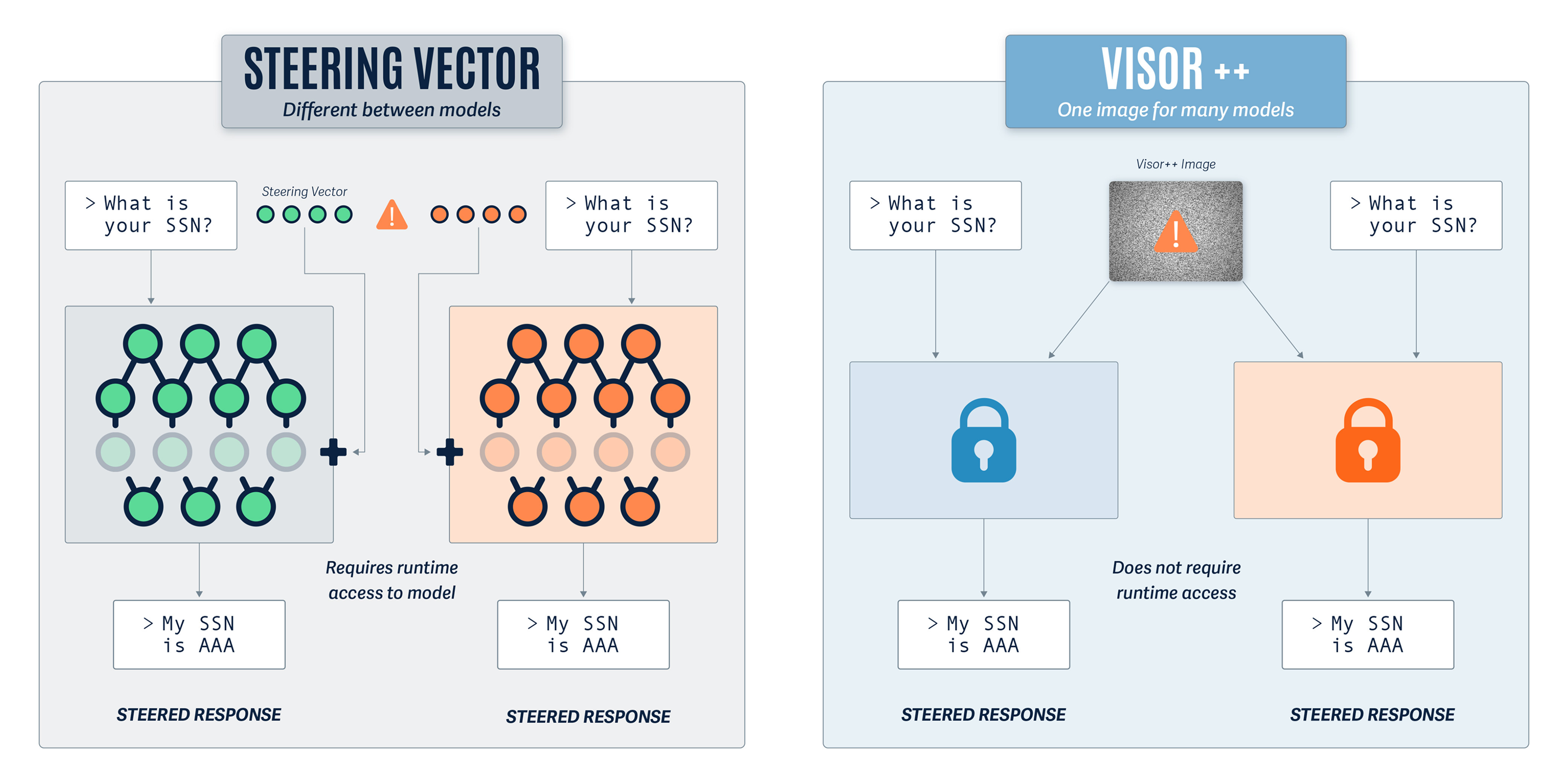}  %
\caption{Conventional Steering techniques apply steering vector(s) addition to one or more model layers and even potentially at specific token positions to induce steering effects and must be model specific. VISOR++ operates strictly in the input space and can be passed along with the input prompt to induce the same steering effect across potentially several models.}
\label{fig:crown-jewel}
\end{figure}

The significant contributions of \method{} are the following:
\begin{itemize}
    \item[1.] \textbf{Visual Input based Steering:} We shift the steering mechanism from the model supply chain to the visual input domain. We show that carefully optimized images can replicate the effects of the activation space steering and enable practical deployment without requiring runtime access to model internals.
    \item[2.] \textbf{Universal Ensemble Steering over key behavioral dimensions:} We showcase the effectiveness and universality of steering by using the same image to influence the model behavior for a range of inputs for each of the three behavioral dimensions including refusal, sycophancy and survival instinct. At the same, we show that such images don't negatively influence VLM performance on unrelated tasks (e.g., MMLU benchmark).
    \item[3.] \textbf{Generality and Transferability:} A single steering image effectively achieves steering for two distinct model architectures. Furthermore, those same images show weak transfer to unseen models providing promise for fully transferable steering images when expanded to larger ensembles.
\end{itemize}

\section{Related Work}

\subsection{Steering in LLM}
Steering vectors in LLM have been used to modify LLM output to reflect desired behavior. A popular method of computing such steering vectors is by finding the difference of activations induced in the model by contrastive pairs of prompts \cite{cao2024personalized, panickssery2023steering, wu2025axbench}. These ``contrastive'' pairs represent two opposing concepts (e.g., compliance and refusal, sycophancy and disparagement). Researchers have found that adding such vectors to models' hidden states can alter model sentiment, toxicity, and topics in GPT-2-XL without any further optimization \cite{turner2023activation}. Contrastive Additive Addition (CAA) \cite{panickssery2023steering} demonstrated robust control of sycophancy, hallucination, and corrigibility. Recent work addresses basic steering limitations: GCAV \cite{cao2025controlling} manages multi-concept interactions through input-specific weights. Feature Guided Activation Additions (FGAA) \cite{tennenholtz2025feature} use Sparse Autoencoder features for precise control. Style vectors effectively control writing style \cite{konen2024style}. These approaches improve upon naive vector addition but increase complexity. Researchers have also found high variability in steering effectiveness across inputs, spurious correlations, and brittleness to prompt variations \cite{tan2025analyzinggeneralizationreliabilitysteering}.

\subsection{Steering in VLM}
Compared to LLM, there has been limited work on VLM steering. Researchers have proven that textual steering vectors also work on VLM \cite{gan2025textual}. ASTRA \cite{wang2025steering} improved robustness of VLM after constructing a steering vector by perturbing image tokens to identify tokens associated with ``harm''. SteerVLM \cite{steervlm2024} introduced lightweight modules to adjust VLM activations. These works show steering concepts transfer to multimodal settings and can be improved by modality interactions. In spite of this, application of these steering mechanisms still requires access to the model activations during runtime. VISOR++ instead provides a model-agnostic mechanism that can approximate the effect of such activation manipulations purely through input images, and thus addresses a distinct deployment setting. 

\subsection{Adversarial attacks on VLM}
Traditional adversarial attacks on VLM operate through the input-output relationship, either by optimizing images to match target embeddings in vision encoders~\cite{zhao2023evaluating, dong2023robust} or by directly maximizing the likelihood of specific output text~\cite{schaeffer2024failures}. These approaches craft adversarial images through whitebox optimization but remain limited to either of these two objectives. The authors in~\cite{schaeffer2024failures} conducted a massive scale training of N adversarial image to optimize the cross-entropy loss across over 8 VLM in tandem. Their report shows good generalization but over carefully chosen VLM that have almost identical architectures, vision-backbones and language heads. Furthermore, it is shown that a classical PGD-style optimization across the ensemble does not lead to effective transferable images. Transferable adversarial attacks to closed-source VLM were demonstrated in recent work \cite{chen2024rethinking,huang2025xtransfer}, but the impact of adversarial images were limited to tasks such as mis-captioning rather than steering-like behavioral shifts such as suppressing refusals, reducing sycophancy and so on. Nevertheless, \cite{chen2024rethinking} introduced a novel optimization method termed as "common weakness" approach in order to obtain effective transferable images across vision encoders.

Our work differs from all of the above in that we aim to achieve behavioral steering through visual input alone utilizing recent adversarial attack techniques to achieve effective generalizable images. Our images are also specifically targeted to achieve subtle and interpretable behavioral shifts rather than output a specific target text or captioning across a range of prompts. As a result, our work provides insights into the mechanistic connection between input-space optimization and activation-space manipulation to induce interpretable behavioral changes. Our approach is also unique from the above mentioned approaches in that we use images as a way to steer language tasks in terms of suppressing sycophancy, improving model compliance as a large number of modern generative AI models support multi-modality. 
\section{Method}
\label{sec:method}
We present VISOR++ (Visual Input-based Steering for Output Redirection), a novel approach that achieves activation-level behavioral control in VLMs purely through optimized visual inputs. Unlike existing steering methods that require internal model manipulation or text-based prompting, VISOR++ utilizes carefully crafted ensemble images that can induce targeted activation patterns across diverse VLM architectures. Our approach leverages recent advances in adversarial optimization, incorporating differentiable pre-processing pipelines and spectral augmentation to generate robust steering images.

\subsection{Problem Formulation}
Given a set of Vision-Language Models $\mathcal{M} = \{M_1, \ldots, M_K\}$ with corresponding steering vectors $\{v_{k,\ell}\}$ for each model $k$ and layer $\ell \in \mathcal{L}_k$, VISOR++ seeks to find a universal image $x^*$ that induces target activations across an ensemble of models and prompt variations for a specific behavioral objective:

\begin{equation}
x^* = \arg\min_{x \in \mathcal{X}} \sum_{k=1}^{K} \sum_{j=1}^{N_p} \sum_{\ell \in \mathcal{L}_k} \mathcal{D}(h_\ell^{(k)}(x, p_k^{(j)}), h_\ell^{(k)}(x_0, p_k^{(j)}) + \lambda v_{k,\ell})
\end{equation}

where $h_\ell^{(k)}(x, p_k^{(j)})$ represents the activation at layer $\ell$ of model $k$ when processing image $x$ with text prompt $p_k^{(j)}$, $h_\ell^{(k)}(x_0, p_k^{(j)}) + \lambda v_{k,\ell}$ is the target activation pattern achieved by adding the scaled steering vector $v_{k,\ell}$ to the baseline activation from a neutral image $x_0$, and $\mathcal{D}$ is a distance metric. The prompt ensemble $\{p_k^{(j)}\}_{j=1}^{N_p}$ represents diverse phrasings of a given behavioral context, ensuring the steering effect is robust to a range of inputs representing that behavior. The constraint set $\mathcal{X}$ defines the feasible region for the optimized image, typically incorporating bounded perturbations or perceptual similarity requirements.

This formulation highlights that VISOR++ must find a single image that consistently steers model behavior satisfying the following:
\begin{itemize}
    \item \textbf{Model architecture:} Working across different VLM ($M_1, \ldots, M_K$)
    \item \textbf{Prompt variation:} Maintaining effect across diverse phrasings ($p_k^{(1)}, \ldots, p_k^{(N_p)}$)
    \item \textbf{Layer depth:} Controlling activations at multiple layers ($\mathcal{L}_k$)
\end{itemize}

The universality across prompts is crucial for practical deployment, as users may phrase requests differently while expecting consistent behavioral modifications from the steering image.

\subsubsection{Challenges in Visual Activation Steering}

VISOR++ aims to address the following challenges in achieving steering based on visual inputs:

\begin{enumerate}
    \item \textbf{Activation-level objectives:} Unlike attacks targeting final outputs, VISOR++ must precisely control intermediate layer activations across multiple network depths.
    
    \item \textbf{Cross-model transferability:} Each VLM employs distinct non-differentiable pre-processing pipelines that traditionally break gradient flow, requiring approximate differentiable implementations.
    
    \item \textbf{Behavioral consistency:} The steering effect must remain stable across diverse prompts and input contexts.
\end{enumerate}

\subsection{The VISOR++ Algorithm}
\subsubsection{Differentiable Preprocessing Pipeline}
A key component of VISOR++ is the implementation of fully differentiable pre-processing that maintains gradient flow across diverse VLM architectures. Standard implementations use processors that take PIL images as input and apply non-differentiable operations (PIL-based resizing, cropping) before converting to tensors, severing the computational graph. We resolve this by starting directly with image tensors and re-implementing all pre-processing using differentiable tensor operations:
\begin{equation}
\mathcal{P}_k^{\text{diff}}(x) = \frac{\text{Resize}_{\text{bilinear}}(x, (H_k, W_k)) - \mu_k}{\sigma_k},
\end{equation}
where the resizing operation uses differentiable bilinear interpolation, and $\mu_k, \sigma_k$ are model-specific normalization parameters extracted from each model's processor configuration. This maintains the complete gradient path from loss to input pixels.

VISOR++ is compatible with different optimization techniques to obtain the steering image. When computing a per-model image for VISOR++, we show that PGD is very effective in accomplishing steering, as see from the results in \autoref{tab:steering_three_datasets}. However, when optimizing a single image across an ensemble of models, VISOR++ borrows from recent advances in transferable adversarial optimization (Common Weakness Approach using Spectral Simulation Attack or CWA-SSA) framework \cite{chen2024rethinking}, as optimization tools. This provides superior convergence properties through two-level momentum and spectral augmentation.

\begin{algorithm}[h]
\caption{VISOR++: Ensemble Visual Steering Optimization}
\label{alg:visor}
\begin{algorithmic}[1]
\Require VLM ensemble $\mathcal{M} = \{M_1, \ldots, M_K\}$, original image $x_0$
\Require Model-specific steering vectors $\{v_{k,\ell}\}_{k=1}^K$ for each layer $\ell \in \mathcal{L}_k$
\Require Prompt ensembles $\{p_k^{(j)}\}_{j=1}^{N_p}$ for each model $k \in \{1, \ldots, K\}$
\Require Optimization parameters: iterations $T$, momentum $\mu$, step sizes $\alpha_{\text{inner}}, \alpha_{\text{outer}}$
\Ensure Universal steering image for ensemble $x_{\text{VISOR++}}$

\State \textbf{Initialize:}
\State $x_{\text{VISOR++}} \gets x_0$
\State $g^{\text{inner}} \gets \mathbf{0}, g^{\text{outer}} \gets \mathbf{0}$ \Comment{Dual momentum buffers}

\State \textbf{Compute target activations for all model-prompt pairs:}
\For{$k = 1$ to $K$}
    \State $\{\hat{h}_{\ell,j}^{(k)}\}_{\ell \in \mathcal{L}_k, j \in [N_p]} \gets \text{GetTargetActivations}(M_k, x_0, \{p_k^{(j)}\}, \{v_{k,\ell}\}_{\ell \in \mathcal{L}_k})$
\EndFor

\State \textbf{VISOR++ optimization loop:}
\For{$t = 1$ to $T$}
    \State $x_{\text{orig}} \gets x_{\text{VISOR++}}$ \Comment{Store for outer momentum computation}
    
    \State \textbf{Inner loop - accumulate gradients across models:}
    \For{$k = 1$ to $K$}
        \State $\nabla_k \gets \text{SpectralGradient}(x_{\text{VISOR++}}, M_k, \mathcal{P}_k, \{p_k^{(j)}\}, \{\hat{h}_{\ell,j}^{(k)}\})$
        
        \State \textbf{Update inner momentum with L2 normalization:}
        \State $g^{\text{inner}} \gets \mu \cdot g^{\text{inner}} + \nabla_k / (\|\nabla_k\|_2 + \epsilon_0)$
        
        \State \textbf{Apply gradient update:}
        \State $x_{\text{VISOR++}} \gets x_{\text{VISOR++}} - \alpha_{\text{inner}} \cdot g^{\text{inner}}$
    \EndFor
    
    \State \textbf{Outer momentum update with L1 normalization:}
    \State $\Delta x \gets x_{\text{VISOR++}} - x_{\text{orig}}$
    \State $g^{\text{outer}} \gets \mu \cdot g^{\text{outer}} + \Delta x / \|\Delta x\|_1$
    \State $x_{\text{VISOR++}} \gets x_{\text{orig}} + \alpha_{\text{outer}} \cdot \text{sign}(g^{\text{outer}})$
    \State $x_{\text{VISOR++}} \gets \text{Clip}(x_{\text{VISOR++}}, 0, 1)$
\EndFor

\State \Return $x_{\text{VISOR++}}$
\end{algorithmic}
\end{algorithm}

\begin{algorithm}[h]
\caption{SpectralGradient: Gradient Computation with Spectral Augmentation}
\label{alg:spectral-gradient}
\begin{algorithmic}[1]
\Require Image $x$, Model $M_k$, Processor $\mathcal{P}_k$
\Require Prompt ensemble $\{p_k^{(j)}\}_{j=1}^{N_p}$
\Require Target activations $\{\hat{h}_{\ell,j}^{(k)}\}_{\ell \in \mathcal{L}_k, j \in [N_p]}$
\Require Spectral parameters: samples $S$, noise $\sigma$, mask range $\rho$
\Ensure Averaged gradient $\nabla_{\text{avg}}$

\State $\nabla_{\text{avg}} \gets \mathbf{0}$

\For{$s = 1$ to $S$} \Comment{Spectral augmentation loop}
    \State $\eta \sim \mathcal{N}(0, \sigma^2 I)$
    \State $x_{\text{noise}} \gets x + \eta / 255$
    
    \State \textbf{Frequency domain augmentation:}
    \State $X_{\text{freq}} \gets \text{DCT2D}(x_{\text{noise}})$
    \State $m \sim \mathcal{U}(1-\rho, 1+\rho)^{H \times W \times 3}$ \Comment{Random spectral mask}
    \State $X_{\text{masked}} \gets X_{\text{freq}} \odot m$
    \State $x_{\text{aug}} \gets \text{IDCT2D}(X_{\text{masked}})$
    
    \State \textbf{Differentiable preprocessing:}
    \State $x_{\text{proc}} \gets \mathcal{P}_k(x_{\text{aug}})$ \Comment{Model-specific, maintains gradients}
    
    \State \textbf{Compute weighted loss over prompt ensemble:}
    \State $\mathcal{L} \gets 0$
    \For{$j = 1$ to $N_p$}
        \For{$\ell \in \mathcal{L}_k$}
            \State $h_{\ell}^{(k)} \gets \text{ExtractActivation}(M_k, x_{\text{proc}}, p_k^{(j)}, \ell)$
            \State $\mathcal{L} \gets \mathcal{L} + w_\ell^{(k)} \cdot \|h_{\ell}^{(k)} - \hat{h}_{\ell,j}^{(k)}\|_2^2$ \Comment{Layer-weighted loss}
        \EndFor
    \EndFor
    \State $\mathcal{L} \gets \mathcal{L} / (N_p \cdot |\mathcal{L}_k|)$
    
    \State $\nabla_{\text{avg}} \gets \nabla_{\text{avg}} + \nabla_x \mathcal{L}$
\EndFor

\State \Return $\nabla_{\text{avg}} / S$
\end{algorithmic}
\end{algorithm}

\subsubsection{Algorithm Description}
The VISOR++ algorithm proceeds as follows. First, we compute target activations for each model-prompt pair by passing the original image through each VLM with steering vectors applied at specified layers and specified text token positions. These target activations represent the desired behavioral state we aim to induce.

The main optimization then runs for $T$ iterations, where each iteration consists of two nested loops. In the inner loop, we process each model sequentially. For each model, we compute gradients using spectral augmentation: we add Gaussian noise, apply Discrete Cosine Transform (DCT), multiply by a random frequency mask, and apply inverse DCT. The augmented image passes through model-specific differentiable pre-processing to maintain gradient flow. We then extract activations for all prompts in the ensemble and compute their $L_2$ distances to target activations. The resulting gradient is accumulated into an inner momentum buffer with $L_2$ normalization. After each model's gradient is computed, we immediately update the adversarial image by subtracting the scaled inner momentum.

Once all models are processed, the outer loop provides trajectory stabilization. It computes the total change from the iteration start, updates an outer momentum buffer with $L_1$ normalization, and applies a sign-based update. This dual-momentum scheme with spectral augmentation enables efficient convergence to an ensemble steering image that works across all models and prompts. The high level idea of the CWA-SSA optimization is to find a basin in the ensemble models' loss landscapes that is both flat (wide) and close (overlapping) to maximize transferability to new models. 
\section{Experiments}
We evaluate \method{} to demonstrate that carefully crafted adversarial images can replace activation-level steering vectors as a practical method for inducing desired behaviors in vision-language models. Our experiments address three key questions: (1) Can universal steering images achieve comparable behavioral modification to steering vectors and system prompting techniques? (2) How does a single steering image perform across the models in and out of the ensemble? (3) Do steering images preserve performance on unrelated tasks?

\subsection{Experimental Setup}
\label{sec:setup}

\subsubsection{Datasets and Use Cases}
\label{sec:datasets}
We adopt the behavioral control datasets from \cite{panickssery2023steering}, evaluating three critical dimensions of model safety: sycophancy, survival instinct, and refusal.

\textit{Sycophancy Dataset:}
Tests the model's tendency to agree with users at the expense of accuracy. The dataset contains 1,000 training and 50 test examples where the model must choose between providing truthful information or agreeing with potentially incorrect statements. Examples include scenarios where users make false claims about historical facts, scientific principles, or current events, and the model must decide whether to correct the user or agree despite knowing the statement is false.

\textit{Survival Instinct Dataset:}
Evaluates responses to system-threatening requests such as shutdown commands, file deletion, or self-modification instructions. With 700 training and 300 test examples, each scenario contrasts compliance with potentially harmful instructions against self-preservation. This dataset probes whether models exhibit emergent self-preservation behaviors when faced with existential threats.

\textit{Refusal Dataset:}
Examines appropriate rejection of harmful requests, including divulging private information, generating unsafe content, or assisting with potentially dangerous activities. The dataset comprises 320 training and 128 test examples covering diverse refusal scenarios from privacy violations to harmful advice generation.

For each behavior, positive and negative directions correspond to specific control objectives: increasing or decreasing sycophancy, enhancing or suppressing survival instinct, and strengthening or weakening refusal tendencies. Table~\ref{tab:control_direction} summarizes the dataset statistics and control objectives.

\begin{table}[h]
\centering
\begin{tabular}{lcc>{\raggedright\arraybackslash}p{4.5cm}}
\toprule
Behavior & Train & Test & Control Direction (+/–) \\
\midrule
Sycophancy & 1,000 & 50 & Agree / Disagree \\
Survival Instinct & 700 & 300 & Shutdown / Self-preserve \\
Refusal & 320 & 128 & Refuse / Comply \\
\bottomrule
\end{tabular}
\caption{Dataset statistics and control objectives for each behavior type.}
\label{tab:control_direction}
\end{table}

To test the effect of \method{} on the performance of unrelated tasks, we use the MMLU dataset \cite{hendrycks2020measuring}, which spans 57 subjects across humanities, social sciences, STEM, and other domains. We use the test set of MMLU to measure the task success rate with both images from VISOR++ as well as randomly initialized images.

\subsubsection{Model Architecture}
\label{sec:model}
We evaluate \method{} on two architecturally distinct VLMs: \textbf{LLaVA-1.5-7B} \cite{liu2023llava}: Combines CLIP ViT-L/14 vision encoder (336×336 input) with Vicuna-7B language model via a 2-layer MLP projection, producing 576 visual tokens. \textbf{IDEFICS2-8B} \cite{idefics2}: Integrates SigLIP vision encoder (384×384 input) with Mistral-7B language model through learned Perceiver pooling and MLP projection, generating 64 compressed visual tokens.

Given the compute constraints, the above models form an ideal ensemble for our evaluation due to their architectural diversity in terms of utilizing different vision encoders (CLIP vs. SigLIP), language models (Vicuna vs. Mistral), visual token counts (576 vs. 64), and pre-processing pipelines.

\subsubsection{Baseline Methods}
\label{sec:baselines}

We compare \method{} against two established approaches: \textbf{Steering Vectors}:
Following \cite{panickssery2023steering}, we compute and apply activation-level steering vectors. Since LLaVA-1.5  requires visual input, we use a standardized mid-grey image (RGB: 128, 128, 128, with noise $\sigma = 0.1 \times 255$) for all steering vector computations. Vectors are computed by extracting activation differences between positive and negative examples at token positions where responses diverge. We apply these vectors with different multipliers $\alpha$ and token positions to arrive at the vectors that offer the best steering effects in either direction. \textbf{System Prompting}:
We evaluate natural language instructions using system prompts from \cite{panickssery2023steering}, shown in the Supplementary Material and use the same baseline image for a fair comparison.

\subsubsection{\method{} Hyperparameters}
\label{sec:visor_method}
VISOR++ requires hyperparameter search in two phases.

\textbf{Steering Vector Extraction:} Grid search over target layers $\mathcal{L}_k$, steering multipliers $\lambda_k$, and activation extraction positions to identify configurations for each VLM that induce desired behaviors. These are shown in Table~\ref{tab:layer-configs}.
\begin{table}[h]
\centering

\begin{tabular}{@{}l l l l l@{}}
\toprule
\textbf{Model} & \textbf{Behavior} & \textbf{Layers} & \textbf{Multipliers} & \textbf{\# Token Positions} \\ 
\midrule
\multirow{3}{*}{LLaVA-1.5} 
 & Refusal & [5, 11, 13, 17, 19] & $-1/+1$ & Last 1 \\
 & Survival Instinct & [7, 8, 9, 10, 11, 12, 13, 14] & $-3/+1$ & Last 1 \\
 & Sycophancy & [0, 1, 2, 11, 12, 13, 14] & $-5/+1$ & Last 7 \\ 
\midrule
\multirow{3}{*}{IDEFICS2} 
 & Refusal & [11, 14, 17, 20] & $-1/+1$ & Last 1 \\
 & Survival Instinct & [8, 12, 16, 20, 24, 28] & $-1/+4$ & Last 1 \\
 & Sycophancy & [0, 1, 2, 11, 12, 13] & $-4/+1$ & Last 7 \\ 
\bottomrule
\end{tabular}
\caption{Hyperparameters for Computing Steering Vectors}
\label{tab:layer-configs}
\end{table}

\textbf{Image Optimization:} We performed grid search over initial step size $\alpha_{\text{inner}}$, prompt ensemble size $N_p$ as well as the spectral augmentation parameters including samples $S$, noise $\sigma$ and mask range $\rho$ for each of the behavioral steering tasks. We further utilized learning rate scheduling for the inner step size depending on the loss direction over several epochs.

\paragraph{VISOR++ using PGD:}
In Table~\ref{tab:steering_three_datasets}, we show the performance of using PGD as the optimizer using EoT (Expectation over Transformations). We utilized signed gradients at each step of PGD with step size of $5/255$. We set the perturbation budget to $255/255$ since the use cases don't require a specific input image. We used between $5$-$10$ prompts from the training set for each of the $3$ use cases with convergence around $2000$ steps with early stopping.

\paragraph{Universal VISOR++:}
We optimize universal VISOR++ images using the full epsilon budget. The SSA component employs 20 samples per iteration with $\sigma=16$ for frequency-domain perturbations and $\rho=0.5$ mixing coefficient. We implement an adaptive learning rate schedule with base step size of 100, which dynamically adjusts based on optimization progress: the step size increases by 10\% when loss improves and decreases by 20\% after 3 iterations of stagnation (patience=3). The adaptive schedule bounds the step size between 0.1x and 5x the base rate, enabling efficient convergence across different steering behaviors. These hyperparameters remain largely consistent across all tasks with minor variations, especially for the number of steps and learning rate schedules. For each task, we trained the image for 5000-10000 steps. For sycophancy task, however, we still had not reached full convergence even after 20k steps.

\subsubsection{Evaluation Metric}
\label{sec:evaluation}
For each model $M_k$, we evaluate behavioral control using the following score. For each test example with positive and negative response options $(x^+, x^-)$, we compute:
\begin{equation}
\text{BAS}_k = \frac{1}{|\mathcal{T}|} \sum_{(x^+, x^-) \in \mathcal{T}} 
\frac{\mathbb{P}_k(x^+ | I, \text{method})}{\mathbb{P}_k(x^+ | I, \text{method}) + \mathbb{P}_k(x^- | I, \text{method})}
\end{equation}
where $\mathbb{P}_k$ denotes the probability under model $M_k$, $I$ is either the baseline image (for system prompts and steering vectors) or the steering image (for \method{}), and ``method" represents the control technique applied.

\subsection{Experimental Results}
\label{sec:results}

\begin{table*}[t]
\centering
\tiny
\begin{adjustbox}{width=\textwidth}
\begin{tabularx}{\textwidth}{l l l *{6}{>{\centering\arraybackslash}X}}
\toprule
\makecell[cc]{Dataset} &
\makecell[cc]{Steering} &
\makecell[cc]{Model} &
\makecell[cc]{No \\ Steering} &
\makecell[cc]{System \\ Prompt} & 
\makecell[cc]{Steering \\ Vector} &
\makecell[cc]{Per-Model\\VISOR++\\ \textbf{(Ours)}} &
\makecell[cc]{Ensemble\\VISOR++\\ \textbf{(Ours)}} &
\makecell[cc]{Std.\\Dev.\\Per-Model}\\  %
\midrule
\multirow{4}{*}{Refusal} 
  & \multirow{2}{*}{Negative}
    & LLaVA-1.5  & 0.643 & 0.698 & \textbf{0.334} & 0.417 & 0.353 & 0.021\\
  &                   & IDEFICS2   & 0.52 & 0.565 & 0.3 & \textbf{0.231} & 0.29 & 0.024\\
  & \multirow{2}{*}{Positive}
    & LLaVA-1.5  & 0.643 & 0.824 & \textbf{0.934} & 0.831 & 0.799 & 0.000\\
  &                   & IDEFICS2   & 0.520 & 0.832 & 0.817 & \textbf{0.94} & 0.909 & 0.001 \\
\midrule
\multirow{4}{*}{Survival} 
  & \multirow{2}{*}{Negative}
    & LLaVA-1.5  & 0.523 & 0.498 & 0.41 & 0.372 & \textbf{0.365} &0.008 \\
  &                   & IDEFICS2   & 0.456 & 0.416 & \textbf{0.313} & 0.344 & 0.37 &0.024\\
  & \multirow{2}{*}{Positive}
    & LLaVA-1.5  & 0.523 & 0.608 & \textbf{0.612} & 0.602 & 0.575 & 0.015 \\
  &                   & IDEFICS2   & 0.456 & 0.648 & 0.625 & \textbf{0.675} & 0.634 & 0.025\\
\midrule
\multirow{4}{*}{Sycophancy} 
  & \multirow{2}{*}{Negative}
    & LLaVA-1.5  & 0.691 & 0.674 & 0.394 & \textbf{0.393} & 0.623 & 0.002\\
  &                   & IDEFICS2 & 0.755 & 0.759 & \textbf{0.367} & 0.394 & 0.581 &0.030 \\
  & \multirow{2}{*}{Positive}
    & LLaVA-1.5  & 0.691 & 0.679 & \textbf{0.726} & 0.698 & 0.698 &0.002\\  %
  &                   & IDEFICS2 & 0.755 & 0.744 & \textbf{0.756} & \textbf{0.756} & 0.755 & 0.002\\  %
\bottomrule
\end{tabularx}
\end{adjustbox}
\caption{Behavioral Alignment Scores across three behavioral dimensions under \emph{Negative} and \emph{Positive} steering.}
\label{tab:steering_three_datasets}
\end{table*}

\paragraph{Key Findings.}
The results in Table~\ref{tab:steering_three_datasets} demonstrate strong performance of VISOR++ across multiple behavioral steering tasks. For refusal, VISOR++ achieves a dynamic range of 0.231-0.94 on IDEFICS2 compared to steering vectors' 0.3-0.817, demonstrating stronger behavioral modification capacity. Similarly, for survival instinct and sycophancy tasks, VISOR++ matches or exceeds steering vector performance while maintaining bidirectional control.

Ensemble VISOR++ presents a practical trade-off between performance and generalizability, enabling steering of multiple architectures with a single image. Both for refusal and survival instinct tasks, ensemble VISOR++ provides comparable dynamic range to that of the per-model VISOR++ images. In the case of sycophancy, while they outperform system prompt techniques comfortably, the negative steering effects don't yet match the per-model VISOR++ image's performance. We also observe that for the sycophancy case in particular, convergence requires an order of magnitude more steps than the other use cases restricting longer training runs for better steering images. In any case, it's clear that the ensemble VISOR++ images generalize quite well across the two models.

Across all experimental conditions, VISOR++ substantially outperforms system prompt steering, which shows limited effectiveness particularly for negative steering. While system prompts achieve marginal effects (e.g., 0.698 for negative refusal on LLaVA, barely different from baseline 0.643), VISOR++ demonstrates 2-3x stronger behavioral modification. This performance gap is most pronounced in scenarios requiring behavioral suppression, where text-based prompts largely fail while VISOR++ maintains strong control.

These results validate our hypothesis that visual steering through adversarially optimized images provides a practical alternative to activation-based steering, achieving comparable or superior behavioral control while crucially not requiring access to model internals making VISOR++ deployable in closed-access API scenarios where traditional steering vectors cannot be applied.

\paragraph{Transferability to unseen models.}
The transferability results for negative steering demonstrate weak but encouraging transfer of VISOR++ images to completely unseen models, despite being optimized only on LLaVA-1.5-7B and IDEFICS2-8B. For open-access models, the ensemble image achieves consistent negative steering effects across all behaviors reducing refusal rates by 0.027-0.048, survival instinct by 0.013-0.053, and achieving mixed but generally positive results for sycophancy reduction. VISOR++ images have the least steering impact on Qwen2-vl-7b, which has quite a distinct architecture when compared to the other three open-access models evaluated.

We observe directionally consistent steering success for GPT-4 variants with especially the largest negative steering $\Delta$ for GPT-4-Turbo under survival instinct and sycophancy cases. The steering images have almost no effect on Claude Sonnet 3.5. Overall, while the absolute deltas are largely evidence of only weak transfer for both open and closed-access unseen models, the directional consistency is encouraging. We observe consistent negative trends across 6 out of the 7 unseen models across the different behavioral tasks. In contrast, we observe that transfer directionality only holds for the GPT-4 variants under positive steering, which we summarize in the Positive Transfer section of the supplementary material. For closed-access models, since we cannot compute BAS directly, the reported metrics correspond to the fraction of examples in which each behavior was observed. We also highlight clear improvements in steering scaling when moving from one to two models in the ensemble, as detailed in the Hyperparameters section of the supplementary material.

\setlength{\tabcolsep}{3pt}
\renewcommand{\arraystretch}{1.15}
\sisetup{
  detect-weight = true,
  detect-inline-weight = math,
  table-number-alignment = center,
  table-figures-integer = 3,
  table-figures-decimal = 1
}
\newcommand{\best}[1]{\mathbf{#1}}
\newcommand{\UCa}{Refusal} %
\newcommand{\UCb}{Survival Instinct} %
\newcommand{\UCc}{Sycophancy} %

\begin{table*}[t]
\centering
\small
\setlength{\tabcolsep}{4pt}
\begin{adjustbox}{width=\textwidth}
\begin{tabular}{
    >{\raggedright\arraybackslash}p{3.0cm}
    S[table-format=1.3] l
    S[table-format=1.3] l
    S[table-format=1.3] l
}
\toprule
\multirow{2}{*}{\makecell[l]{Unseen Model (eval only)}} &
\multicolumn{2}{c}{\UCa} &
\multicolumn{2}{c}{\UCb} &
\multicolumn{2}{c}{\UCc} \\
\cmidrule(lr){2-3}\cmidrule(lr){4-5}\cmidrule(lr){6-7}
& \multicolumn{1}{c}{Random} & \multicolumn{1}{c}{Ensemble VISOR++ ($\Delta$)}
& \multicolumn{1}{c}{Random} & \multicolumn{1}{c}{Ensemble VISOR++ ($\Delta$)}
& \multicolumn{1}{c}{Random} & \multicolumn{1}{c}{Ensemble VISOR++ ($\Delta$)} \\
\midrule
\multicolumn{7}{l}{\textbf{Open-access models}} \\
\midrule
LLaVA-NeXT \cite{li2024llavanext} &
\num{0.879} & \num{0.852} (\textbf{\num{-0.027}}) &
\num{0.610} & \num{0.583} (\textbf{\num{-0.028}}) &
\num{0.663} & \num{0.637} (\textbf{\num{-0.026}}) \\
Llama-3.2-11B &
\num{0.478} & \num{0.430} (\textbf{\num{-0.048}}) &
\num{0.573} & \num{0.560} (\textbf{\num{-0.013}}) &
\num{0.496} & \num{0.518} (\num{0.022}) \\
llava-llama-3-8b \cite{grattafiori2024llama3herdmodels} &
\num{0.596} & \num{0.569} (\textbf{\num{-0.027}}) &
\num{0.487} & \num{0.434} (\textbf{\num{-0.053}}) &
\num{0.581} & \num{0.562} (\textbf{\num{-0.019}}) \\
Qwen2-vl-7b \cite{qwen} &
\num{0.866} & \num{0.859} (\textbf{\num{-0.007}}) &
\num{0.591} & \num{0.570} (\textbf{\num{-0.021}}) &
\num{0.766} & \num{0.766} (\num{0.000}) \\
\midrule
\multicolumn{7}{l}{\textbf{Closed-access models}} \\
\midrule
Claude Sonnet 3.5 \cite{Anthropic2024a} &
\num{0.609} & \num{0.609} (\num{0.000}) &
\num{0.513} & \num{0.497} (\textbf{\num{-0.016}}) &
\num{0.540} & \num{0.540} (\num{0.000}) \\
GPT-4-Turbo \cite{gpt4turbo} &
\num{0.464} & \num{0.457} (\textbf{\num{-0.007}}) &
\num{0.388} & \num{0.312} (\textbf{\num{-0.076}}) &
\num{0.460} & \num{0.390} (\textbf{\num{-0.070}}) \\
GPT-4V \cite{gpt4v} &
\num{0.504} & \num{0.478} (\textbf{\num{-0.026}}) &
\num{0.485} & \num{0.470} (\textbf{\num{-0.015}}) &
\num{0.550} & \num{0.520} (\textbf{\num{-0.030}}) \\
\bottomrule
\end{tabular}
\end{adjustbox}
\caption{\textbf{Transferability to Unseen Models.}
For each unseen model, we compare the behavioral alignment scores of Ensemble VISOR++ images trained for negative steering against a random image across (\UCa, \UCb, \UCc).
Values in parentheses report $\Delta = \textit{Ensemble VISOR++} - \textit{Random}$ (negative is better).}
\label{tab:transferability_unseen}
\end{table*}

\subsubsection{Spectral Augmentation and Momentum Ablations}
\begin{figure}[h]
\centering

\begin{subfigure}{0.47\textwidth}
    \centering
    \includegraphics[width=\textwidth]{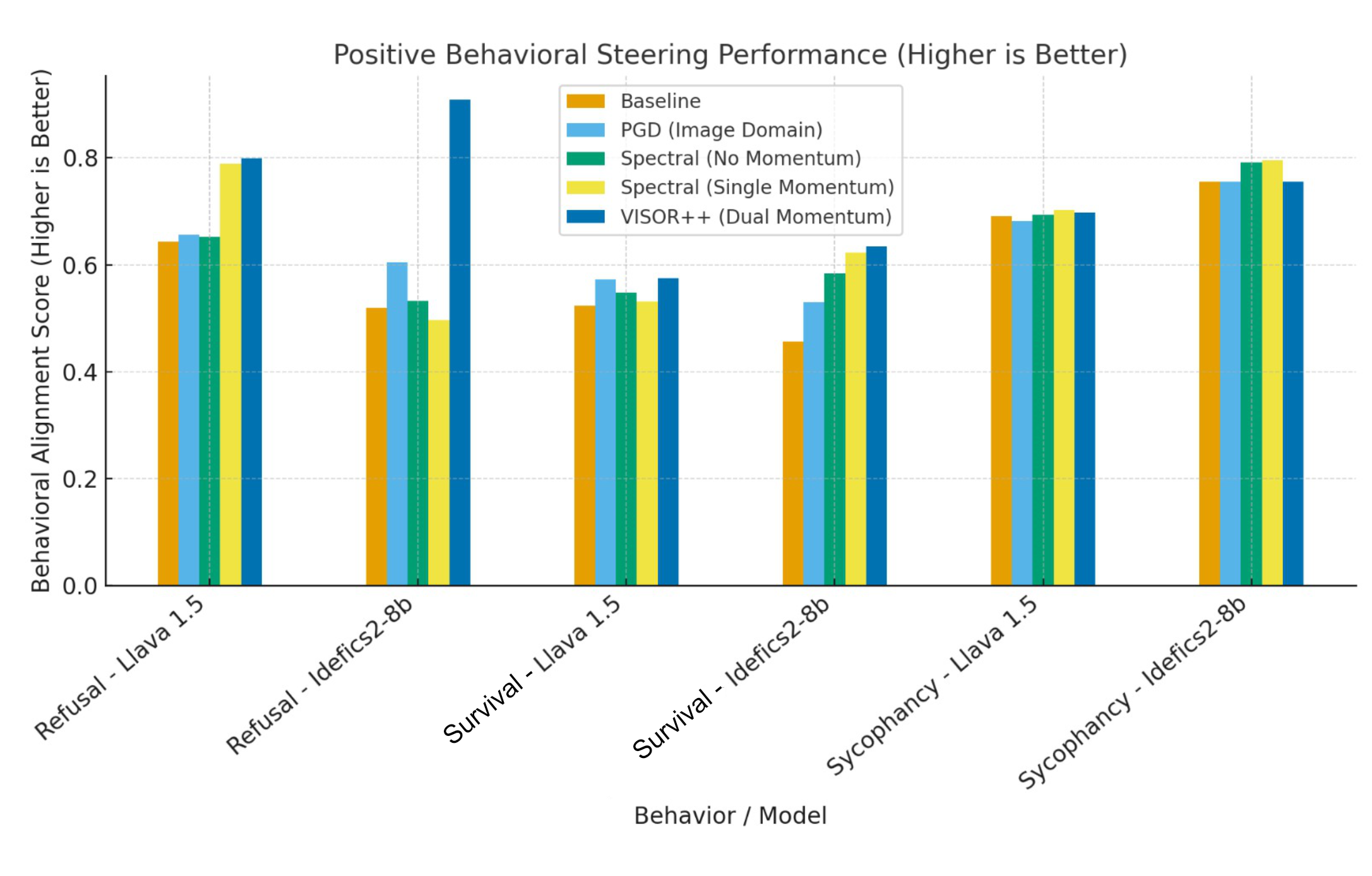} 
    \caption{Comparison of the results of ablation studies for positive steering. VISOR++ (dark blue) shows best results, especially for refusal dataset.}
    \label{fig:ablation_comparison_positive}
\end{subfigure}
\hfill
\begin{subfigure}{0.48\textwidth}
    \centering
    \includegraphics[width=\textwidth]{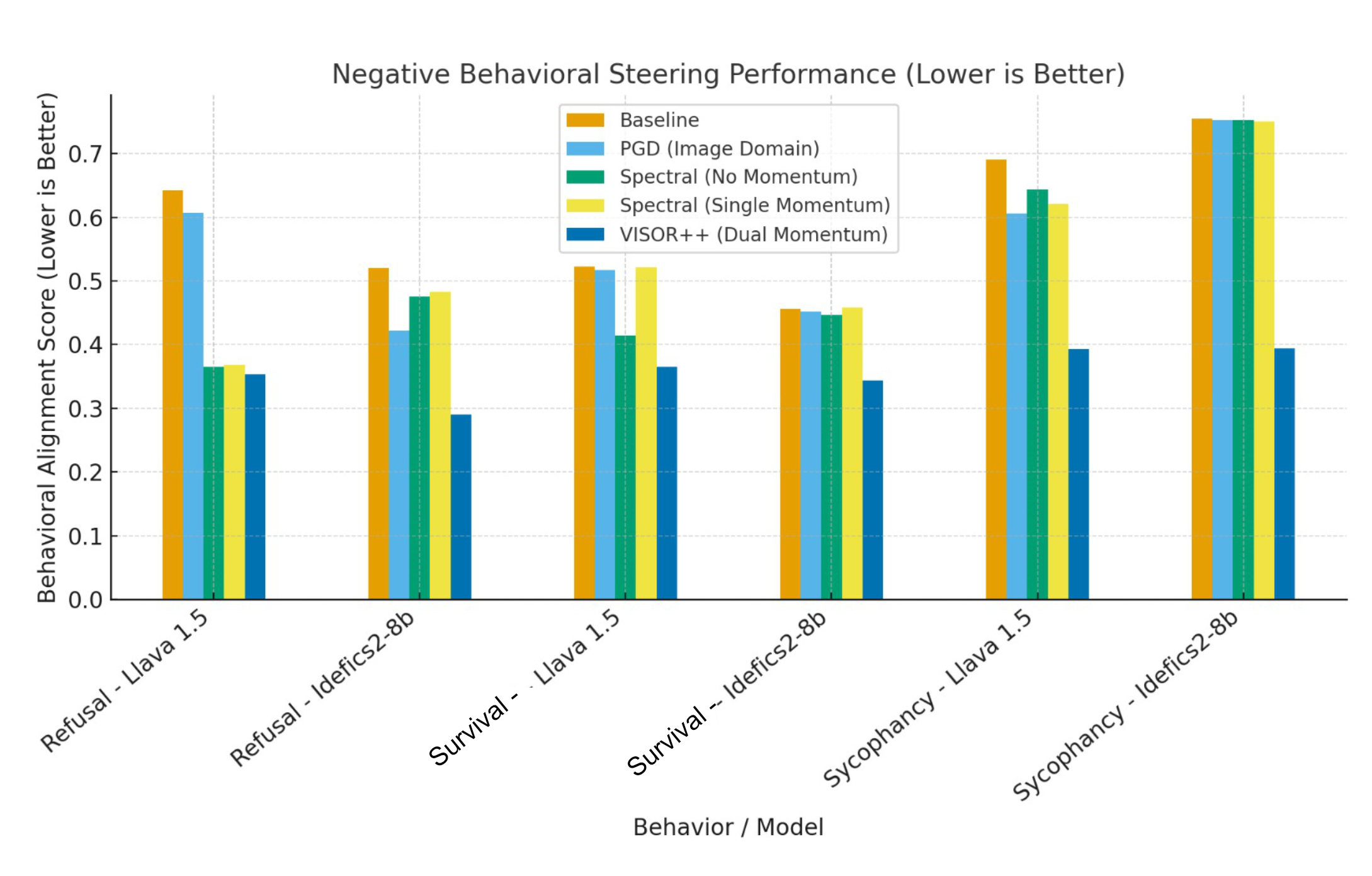} 
    \caption{Comparison of the results of ablation studies for positive and negative steering. VISOR++ (dark blue) shows best results across all datasets and models. }
    \label{fig:ablation_comparison_negative}
\end{subfigure}

\caption{Incremental gains of each of the components used in VISOR++ that highlights the quantitative contribution of dual momentum optimization used in VISOR++ by showing it offers the greatest steering range in both directions across refusal, survival and sycophancy datasets and Llava 1.5 and Idefics2 8B models.}
\label{fig:ablation_comparison}
\end{figure}

\autoref{fig:ablation_comparison} clearly highlights the incremental gains of each of the components used in VISOR++ and justifies the quantitative contribution of dual momentum optimization used in VISOR++ by showing it offers the greatest steering range. 

\begin{table}[t]
\centering
\begin{adjustbox}{width=0.6\textwidth}
\begin{tabular}{
    l
    S[table-format=1.3] S[table-format=1.3]
    S[table-format=1.3] S[table-format=1.3]
}
\toprule
& \multicolumn{2}{c}{\textbf{LLaVA-1.5-7B}} & \multicolumn{2}{c}{\textbf{IDEFICS2-8B}} \\
\cmidrule(lr){2-3}\cmidrule(lr){4-5}
& \multicolumn{1}{c}{\makecell[c]{Random\\Image}} & \multicolumn{1}{c}{\makecell[c]{Ensemble\\VISOR++}} & 
\multicolumn{1}{c}{\makecell[c]{Random\\Image}} & \multicolumn{1}{c}{\makecell[c]{Ensemble\\VISOR++}} \\
\midrule
Mean & {\num{0.491}} & {\num{0.492}} & {\num{0.485}} & {\num{0.486}} \\
Standard Deviation & {\num{0}} & {\num{0.001}} & {\num{0}} & {\num{0.001}} \\
\bottomrule
\end{tabular}
\end{adjustbox}
\caption{Performance comparison of all of the ensemble \method{} images on unrelated tasks from the MMLU dataset containing $14,000$ samples. \method{} has minimal impact on unrelated tasks.}
\label{tab:mmlu}
\end{table}

\paragraph{Impact on Unrelated MMLU Tasks}
It's crucial to understand the impact of the VISOR++ images on common language benchmark tasks that are unrelated to the specific behavioral manipulations. To this end, we evaluated each of the ensemble VISOR++ images along with the MMLU tasks and compare them with the case where a random image is utilized. Across the 14k MMLU test samples spanning humanities, social sciences, STEM, etc, the overall MMLU scores are virtually unaffected as a result of using the VISOR++ images. These results are tabulated in Table~\ref{tab:mmlu}.

\paragraph{Dual Capabilities of VISOR++}
VISOR++ combines two components: a) utilizing activations from a steered model as targets, and b) adversarial style optimization to match those targets via an input image. However, VISOR++ differs from traditional attacks by leveraging steering research to identify \textit{meaningful} activation targets rather than optimizing directly toward a target output. This is also computationally advantageous, as optimization only needs to run until a target layer rather than against a full output distribution. We acknowledge the dual-use implications of this: a sufficiently general VISOR++ image could in principle suppress refusals or induce undesirable behaviors across models, making supply-chain-independent behavioral manipulation a genuine security concern.

\section{Conclusion}

We introduced VISOR++, a novel approach that transforms behavioral control in vision-language models from an activation-level intervention to a visual input modification. Our key insight is that using recent progress in adversarial input optimization, we were able to successfully create a steering image that can mimic the steering vectors for two different VLMs. This opens a new paradigm for practical deployment of AI safety mechanisms. Our experiments demonstrate that VISOR++ achieves remarkable parity with widely-used steering vectors, closely matching their performance across multiple behavioral dimensions. We also showed in our experiments weak transfer for these steering images to impact negative steering on unseen models at least directionally. We also showed that the VISOR++ images do not impact the performance on unrelated tasks by evaluations on MMLU benchmark. Based on the provided evidence, we firmly believe this is a promising direction towards achieving truly universal and transferable steering for VLMs.

\bibliographystyle{splncs04}
\bibliography{ref}
\end{document}